\documentclass{article}

 \usepackage[dblblindworkshop, final]{neurips_2025}

\usepackage[utf8]{inputenc} 
\usepackage[T1]{fontenc}    
\usepackage{hyperref}       
\usepackage{url}            
\usepackage{booktabs}       
\usepackage{amsfonts}       
\usepackage{nicefrac}       
\usepackage{microtype}      
\usepackage{xcolor}         
\usepackage{graphicx}
\usepackage{amsmath}
\usepackage{multirow}
\usepackage{wrapfig}

\title{Estimating Clinical Lab Test Result Trajectories from PPG using Physiological Foundation Model and Patient-Aware State Space Model – a UNIPHY+ Approach}

\author{%
  Minxiao Wang \quad Runze Yan \quad Carol Li \quad Saurabh Kataria \quad Xiao Hu \\
  School of Nursing, Emory University, Atlanta, GA \\
  \texttt{\{mwang80,xiao.hu\}@emory.edu} \\
  \And
  Matthew Clark \quad Timothy Ruchti \\
  NKDHS, Inc \\ Irvine, CA \\
  \And
  Timothy G. Buchman \quad Sivasubramanium V Bhavani \\
  Emory Healthcare \\ Atlanta, GA \\
  \And
  Randall J. Lee \\
  UCSF \\ San Francisco, CA \\
}

\begin{document}

\maketitle

\begin{abstract}
Clinical laboratory tests provide essential biochemical measurements for diagnosis and treatment, but are limited by intermittent and invasive sampling. In contrast, photoplethysmogram (PPG) is a non-invasive, continuously recorded signal in intensive care units (ICUs) that reflects cardiovascular dynamics and can serve as a proxy for latent physiological changes. We propose UNIPHY+Lab, a framework that combines a large-scale PPG foundation model for local waveform encoding with a patient-aware Mamba model for long-range temporal modeling. Our architecture addresses three challenges: (1) capturing extended temporal trends in laboratory values, (2) accounting for patient-specific baseline variation via FiLM-modulated initial states, and (3) performing multi-task estimation for interrelated biomarkers. We evaluate our method on the two ICU datasets for predicting the five key laboratory tests. The results show substantial improvements over the LSTM and carry-forward baselines in MAE, RMSE, and $R^2$ among most of the estimation targets. This work demonstrates the feasibility of continuous, personalized lab value estimation from routine PPG monitoring, offering a pathway toward non-invasive biochemical surveillance in critical care.
\end{abstract}

\section{Introduction}

Clinical laboratory tests are fundamental to modern medicine, providing quantitative measures that guide diagnosis, risk stratification, and treatment~\cite{takieddine2025laboratory}. Core components such as electrolytes, lactate, blood glucose, and acid–base markers offer critical insights into a patient's metabolic, renal, and perfusion status~\cite{lab1}. However, these tests are inherently intermittent and invasive~\cite{lab2}, typically requiring venous or arterial blood draws. Arterial sampling for acid–base analysis is particularly painful, technically challenging, and requires trained personnel~\cite{abg}, while samples must be processed immediately under strict handling conditions. Even in the intensive care unit (ICU), where labs are ordered more frequently, the temporal sparsity of these measurements makes it difficult to capture rapid changes in physiological parameters related to blood flow, metabolism, or acid–base balance~\cite{ICU}. In contrast, ICU monitors continuously collect high-frequency, non-invasive physiological signals such as photoplethysmography (PPG), which reflects cardiovascular dynamics and is automatically acquired via bedside sensors. Because PPG is already widely available and recorded without additional effort or discomfort, it presents an opportunity to estimate hidden physiological changes~\cite{tazarv2021deep}.

Recent studies have demonstrated the utility of deep learning models applied to PPG for disease detection and physiological parameter estimation. Applications include cardiovascular conditions such as hypertension and atrial fibrillation~\cite{elgendi2019use, bashar2019atrial, liao2022impact, ding2023photoplethysmography} as well as parameters like heart rate and blood pressure~\cite{tian2025paralleled}. More recently, large-scale PPG foundation models (FMs)~\cite{pulseppg2025, pillai2025papagei, chen2025gptppg, abbaspourazad2023} have shown strong generalization across downstream tasks and early evidence of estimating lab-like variables such as glucose trends. While promising, these capabilities have been explored only in narrow contexts and remain in their early stages, leaving open the question of whether PPG FMs can be extended to formal clinical laboratory test estimation. However, a major barrier is that current PPG FMs operate on short input segments (5–30s), focusing on instantaneous cardiovascular features rather than the long-term temporal patterns needed to model biochemical processes that evolve over minutes to hours. Clinical laboratory tests differ fundamentally from vital signs in that they reflect the underlying metabolic and homeostatic processes, which require extended temporal contexts to estimate~\cite{ppgreview}. To our knowledge, no prior work has combined PPG foundation models with long-range sequence modeling to estimate a standard clinical laboratory panel from continuous waveform data.

To address this gap, we propose UNIPHY+Lab, a framework for predicting clinical laboratory parameters directly from continuous PPG waveforms. Our method combines a PPG-based foundation model for local feature extraction with a state-space model (SSM) that integrates information over extended time windows. We address three key challenges: (1) modeling long-range temporal trends in lab values, (2) accounting for patient-specific baseline variation, and (3) performing multi-task, multi-channel estimation for interrelated labs. We focus on the Basic Metabolic Panel (BMP), a commonly ordered set of tests for monitoring disease progression, and evaluate our approach on two ICU datasets. Our results demonstrate that UNIPHY+Lab can provide high-temporal-resolution lab estimates without invasive sampling, enabling continuous and personalized biochemical monitoring in the ICU.

\begin{wrapfigure}{r}{0.66\linewidth} 
    \centering
    \includegraphics[width=\linewidth]{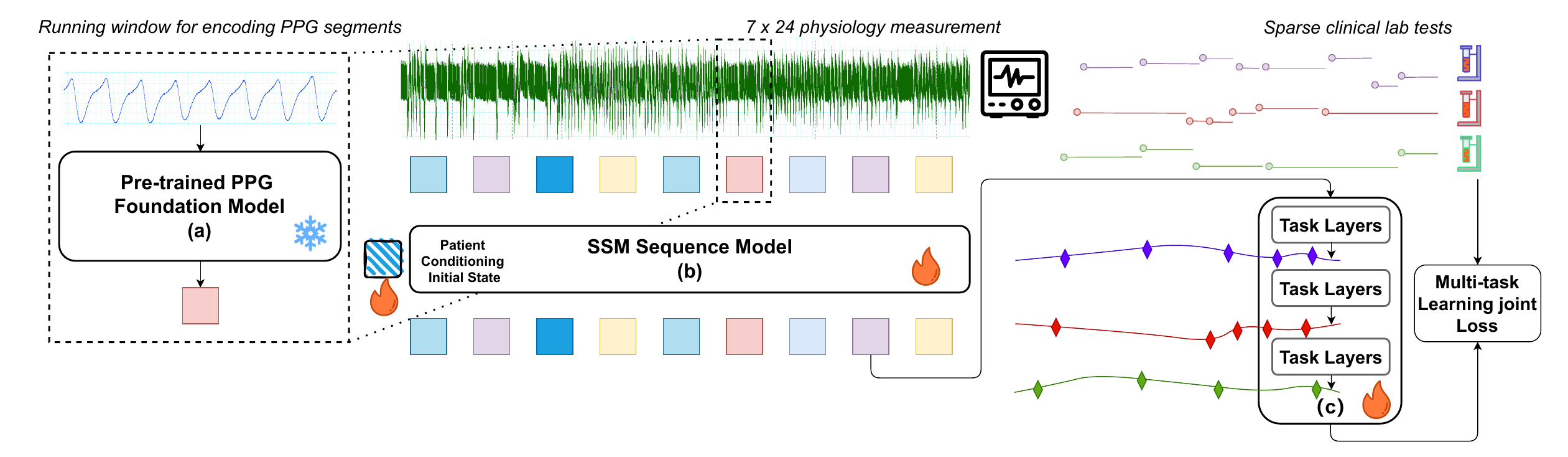}
    \caption{Overview of the UNIPHY+Lab framework. Continuous PPG waveforms are segmented into short windows and encoded by a pretrained foundation model (PPG-GPT) to capture local cardiovascular dynamics. These embeddings are processed by a multi-layer Mamba-based state-space model (SSM) to capture long-range temporal dependencies. Patient-specific conditioning is introduced via a personalized initialization of the SSM state, and multi-task regression heads estimate trajectories for multiple laboratory biomarkers.}
    \label{fig:1}
\end{wrapfigure}

\section{Method}
To address the challenges of long-range physiological modeling, multi-task clinical lab estimation, and patient-specific variability, we propose a novel architecture UNIPHY+Lab, as illustrated in Figure~\ref{fig:1}. UNIPHY+Lab consists of three core components: a foundation model-based local encoder, a patient-aware state-space model for long-range temporal modeling, and a multi-head lab estimation module with task-specific decoding for estimation.

\textbf{Local PPG Feature Encoder via Foundation Model} For the continuous PPG waveform input, we first segment it into fixed length windows (30 seconds in our study) and encode each window using a pretrained transformer-based foundation model. In this work, we adopt the PPG-GPT model~\cite{chen2025gptppg}, which has four different scales: 19 M (selected for this work), 85 M, 345 M, and 1B parameters are pretrained with 200 million 30-second PPG samples. For a window, the encoder splits the 30-second 40 Hz PPG signal into 30 non-overlapping patches as the input sequence of the GPT architecture. Then, we select the output feature of the last patch as the representation embedding for the current encoding window. This encoder captures local morphology and short-range waveform dynamics such as pulse shape, rhythm, and variability. These representations serve as localized features encoding the immediate cardiovascular context. The encoder can be optionally fine-tuned on the downstream task to adapt to the target lab prediction domain.

\textbf{Long-Range Continuous Modeling via SSM} To track latent physiological changes over extended time windows, we employ a stack (default 4 layers) of SSM blocks built upon the Mamba architecture~\cite{gu2023mamba} as the backbone encoder. Each state space block contains two sequential components: a state space layer with a selective scan mechanism that updates temporal memory across the entire sequence, and a gated multilayer perceptron (gMLP) network that enhances nonlinear representations~\cite{ramesh2025lyra}. By maintaining internal hidden states that evolve across time, these SSM blocks enable the model to accumulate long-range dependencies, which are critical for continuous modeling and accurate laboratory value prediction.

\textbf{Patient Conditioning Initial State for SSM} The recent study~\cite{buitragounderstanding} demonstrated that the length generalization of SSM models can benefit from initial state interventions (such as State Passing and Truncated Backpropagation Through Time). The core idea of the initial state intervention is to make the initial state "attainable," thereby it decouples the state position information from the state distribution. 

Inspired by the initial state intervention, we adopt a learnable initial state for the long-range continuous SSM. Meanwhile, we note that physiological waveforms often exhibit individual-specific distributions, making them potentially more personally identifiable than language data (the inputs of the empirical study in~\cite{buitragounderstanding}). Hence, we incorporate a FiLM-like~\cite{perez2018film, turkoglu2022film} patient-specific modulation into the initialization of the SSM state. 

\begin{wrapfigure}{r}{0.4\linewidth} 
    \centering
    \includegraphics[width=\linewidth]{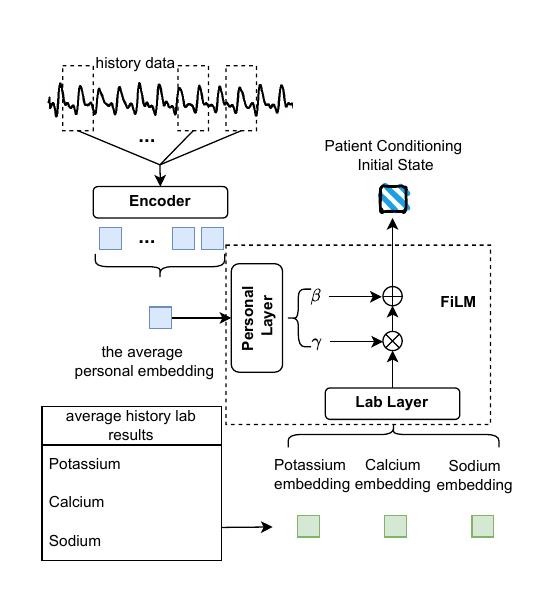}
    \caption{The Patient Conditioning Initial State (PCS) module enables the model to capture patient-specific lab trajectories more effectively.}
    \label{fig:2}
\end{wrapfigure}

As illustrated in Figure~\ref {fig:2}, the patient conditioning initial state (PCS) base is integrated from the embedding for multiple average results from different history labs. The initial state will be modulated by another embedding vector, which is the mean pooling of randomly sampled waveform embeddings of the same patient's history. This initial state will participate in the state transitions, enabling the model to adapt its trajectory dynamics based on the patient's context.

\textbf{Multiple Lab Estimation and Multi-task Learning} To address the heterogeneity of lab values, we adopt a multi-task prediction strategy in which each target biomarker is modeled using a distinct task layer. This design enables the architecture to specialize in both the statistical distribution and temporal dynamics of each biomarker. For each lab $k$, the corresponding task layer includes a lab-specific SSM block and an estimation head. It receives the shared temporal representation from the backbone encoder and outputs a scalar lab value prediction. The lab-specific regression SSM block also includes a PCS module, but conditioned on a single lab’s historical values. During training, we ensure causal estimation by using only features available prior to the target lab measurement. Ground-truth values are linearly interpolated to align with the SSM outputs, and training is guided by a multi-task loss~\cite{liebel2018auxiliary} with uncertainty-based weighting $L_{\text{total}} = \sum_{i=0}^{N} \left( \frac{1}{2\sigma_{i}^2} L_i + \log(\sigma_{i}) \right)$, where $L_i$ denotes the MAE loss for the $i$-th task and $\sigma_{i}$ is a learnable task-specific uncertainty parameter. This formulation allows the model to adaptively balance the contribution of each task according to its estimated uncertainty.

\section{Experiment}

\textbf{Dataset} We used two large-scale ICU datasets containing high-resolution photoplethysmography (PPG) waveforms, continuously recorded vital signs, and structured electronic health records (EHR) with laboratory results and clinical events: an ICU dataset collected in an academic medical institute~\cite{drew2014insights, xiao2020generalizability} (3,796 patients) and the MIMIC-III Waveform Database Matched Subset~\cite{mimic_waveform_matched, johnson2016mimic, goldberger2000physiobank} (4,146 patients). Both datasets provide up to 7 days of temporal coverage across physiological and clinical data streams. Given the PPG FM extract embedding for each 30 seconds, the maximum length of the input for SSM is 20160.

\textbf{Target Biomarkers} To focus on clinically meaningful and routinely measured outcomes, we chose five biomarkers as the estimation targets: Potassium, Calcium, Sodium, Glucose, and Lactate. Electrolytes like potassium, calcium, and sodium affect heart rhythm, muscle contraction, and blood pressure regulation—factors that shape the pulse waveform. Meanwhile, abnormal levels of glucose and lactate can lead to changes in vascular tone and circulation, which are also detectable through PPG. To enable stable training and handle differences in value ranges across lab channels, we apply min-max normalization on each 30-second PPG segment and channel-wise min-max normalization to the ground-truth lab values based on global statistics computed from the training set. 

\textbf{Baseline} We adopted a Long Short-Term Memory (LSTM) model to integrate a fixed-length (default 10 embeddings for 5 minutes) window of embeddings extracted from the same pretrained model, centered around the time of the target measurement. The baseline model was adapted to the multitask setting via hard parameter sharing, where a single backbone network is jointly trained to predict all estimation targets. In addition, we implement a non-parametric last observation carried forward (LOCF) baseline that uses the most recent available lab result as the predicted value. This baseline represents a common clinical reference point, especially for slowly changing or sparsely measured labs.


\section{Results}

As shown in Table~\ref{tab:1}, we report standard regression metrics including mean absolute error (MAE), mean error (ME), root mean squared error (RMSE), and coefficient of determination ($R^2$) on both datasets. We report both ME and RMSE together, as ME quantifies prediction bias while RMSE reflects the variance and overall dispersion of errors. We observe that the baseline LSTM model and UNIPHY+Lab without the PCS have much larger errors and lower $R^2$ coefficients than LOCF. That indicates the baseline model, which relies on short windows of FM embeddings, struggles to capture meaningful relationships between physiological states and lab values with an independent and identically distributed (i.i.d.) assumption on the embeddings and lab values pairs. Although UNIPHY+Lab without PCS achieved higher performance than baseline LSTM, it still suffers from the patient-specific variability problem. 

We find that incorporating the PCS into the Mamba backbone consistently improves performance across both single-task and multi-task settings. Mamba’s selective scan mechanism excels at integrating long-range dependencies, but without personalization, it still learns from population-level patterns that may overlook patient-specific signal baselines and idiosyncratic dynamics. PCS addresses this by initializing the model’s internal state with embeddings derived from each patient’s historical PPG segments and prior lab values, modulated via FiLM to capture individual physiological distributions. This personalized initialization allows the state transitions to adapt immediately to a patient’s unique baseline and variability, enabling the model to more accurately align with true lab trajectories over time. As shown in Table~\ref{tab:1}, adding PCS yields consistent gains in $R^2$ and error metrics, reflecting its ability to preserve long-term temporal coherence while accounting for individual-specific waveform characteristics. 

Interestingly, the results for sodium differ from other biomarkers. LOCF already achieves very strong performance on both datasets, reflecting sodium’s relative temporal stability and slower dynamics in ICU patients. In this case, PCS-enhanced models narrow the gap with LOCF but do not surpass it substantially, suggesting that sophisticated modeling provides less added value for biomarkers with low short-term variability.

We also observe that the multi-task learning provides additional benefits over single-task training. Sharing the backbone representation across related lab prediction tasks encourages the model to leverage inter-lab correlations, improving robustness and reducing overfitting on sparsely measured targets. When combined with PCS, multi-task training consistently yields the highest $R^2$ scores and lowest error metrics in Table~\ref{tab:1}, demonstrating that personalization and multi-task learning are complementary for modeling heterogeneous, patient-specific biochemical trajectories.

\section{Conclusion}

We presented UNIPHY+Lab, a framework that integrates a large-scale PPG foundation model with a patient-aware Mamba state space model for continuous estimation of clinical laboratory values. By combining long-range temporal modeling with a FiLM-modulated patient conditioning initial state, our approach adapts to individual-specific physiological distributions and captures dynamic lab trajectories beyond what population-level models or static baselines can achieve. In addition, the multi-task learning enables the model to exploit correlations among related biomarkers, improving robustness and predictive accuracy for sparsely measured targets. Experiments on the two datasets demonstrate that UNIPHY+Lab substantially outperforms both recurrent and carry-forward baselines in MAE, RMSE, and $R^2$, with the largest gains observed when PCS and multi-task learning are combined. These results highlight the potential of personalized, non-invasive monitoring from routine PPG data to provide high-temporal-resolution biochemical insights in critical care settings. For future work, we see two promising directions for extending this work. First, we will systematically analyze the impact of foundation model scale on downstream performance, exploring how larger or domain-specialized PPG encoders influence generalization and personalization. Second, beyond PPG, we plan to extend UNIPHY+Lab to incorporate additional physiological signals such as ECG, enabling a richer multi-modal representation of cardiovascular state and further improving the fidelity of continuous lab value estimation.

\begin{table}[t]
\centering
\caption{Performance comparison on Institutional and MIMIC-III datasets across different lab estimation tasks, where ST, MT, and PCS refer to single-task, multi-task, and patient conditioning initial state, respectively. Best results are \textbf{bolded}.}
\label{tab:1}
\resizebox{\linewidth}{!}{
\begin{tabular}{llcccccccc}
\toprule
\multirow{2}{*}{Tasks} & \multirow{2}{*}{Methods} & \multicolumn{4}{c}{Institutional} & \multicolumn{4}{c}{MIMIC-III} \\
\cmidrule(lr){3-6} \cmidrule(lr){7-10}
 & & MAE & ME & RMSE & $R^2$ & MAE & ME & RMSE & $R^2$ \\
\midrule
\multirow{5}{*}{Potassium}
 & LSTM          & 0.450 & 0.084 & 0.637 & -0.022 & 0.460 & -0.022 & 0.619 & -0.044 \\
 & LOCF          & 0.376 & -0.015 & 0.659 & -0.046 & 0.118 & \textbf{-0.005} & 0.182 &  0.155 \\
 & Our+MT        & 0.416 & -0.017 & 0.623 &  0.002 & 0.135 & -0.028 & 0.191 &  0.003 \\
 & Our+ST+PCS    & 0.313 & -0.029 & 0.488 &  0.389 & 0.110 & -0.015 & 0.156 &  0.345 \\
 & Our+MT+PCS    & \textbf{0.311} & \textbf{-0.014} & \textbf{0.484} & \textbf{0.397} & \textbf{0.099} & -0.028 & \textbf{0.138} & \textbf{0.460} \\
\midrule
\multirow{5}{*}{Calcium}
 & LSTM          & 0.549 & -0.039 & 0.771 & -0.010 & 0.638 &  0.057 & 0.860 & -0.059 \\
 & LOCF          & 0.296 & -0.010 & 0.477 &  0.587 & \textbf{0.191} & 0.011 & 0.310 &  0.486 \\
 & Our+MT        & 0.549 &  0.016 & 0.727 &  0.004 & 0.319 & -0.017 & 0.432 &  0.024 \\
 & Our+ST+PCS    & 0.287 &  \textbf{0.000} & 0.426 &  0.650 & 0.197 & -0.021 & 0.325 &  0.460 \\
 & Our+MT+PCS    & \textbf{0.284} & -0.004 & \textbf{0.422} & \textbf{0.662} & 0.195 & \textbf{-0.001} & \textbf{0.295} & \textbf{0.613} \\
\midrule
\multirow{5}{*}{Sodium}
 & LSTM          & 3.586 & 0.420 & 4.846 & -0.003 & 4.861 & 0.444 & 6.509 & -0.095 \\
 & LOCF          & \textbf{1.873} & -0.097 & \textbf{2.755} & \textbf{0.742} & \textbf{1.401} & 0.114 & \textbf{2.074} & \textbf{0.731} \\
 & Our+MT        & 3.785 & 0.115 & 5.244 &  0.001 & 3.028 & 0.397 & 3.951 & -0.002 \\
 & Our+ST+PCS    & 2.045 & 0.024 & 2.910 &  0.691 & 1.774 & \textbf{-0.066} & 2.418 &  0.600 \\
 & Our+MT+PCS    & 2.015 & \textbf{-0.002} & 2.863 &  0.702 & 1.737 & 0.232 & 2.389 &  0.658 \\
\midrule
\multirow{5}{*}{Glucose}
 & LSTM          & 39.366 & -5.826 & 64.816 & -0.009 & 36.220 & -3.598 & 57.012 & -0.024 \\
 & LOCF          & 36.143 & \textbf{-3.006} & 77.351 & -0.188 & 13.693 & \textbf{-0.800} & 28.113 & -0.145 \\
 & Our+MT        & 39.321 & -13.450 & 69.252 & -0.029 & 13.899 & -5.165 & 25.350 & -0.038 \\
 & Our+ST+PCS    & 31.917 & -6.009 & 62.047 &  0.230 & 12.254 & -4.534 & 20.705 &  0.230 \\
 & Our+MT+PCS    & \textbf{31.926} & -5.940 & \textbf{59.538} & \textbf{0.240} & \textbf{11.044} & -2.927 & \textbf{19.380} & \textbf{0.285} \\
\midrule
\multirow{5}{*}{Lactate}
 & LSTM          & 1.781 & -0.203 & 3.146 &  0.009 & 1.243 &  0.335 & 1.960 &  0.071 \\
 & LOCF          & 0.736 & -0.180 & 1.318 &  0.505 & \textbf{0.542} & -0.152 & \textbf{0.905} & \textbf{0.577} \\
 & Our+MT        & 1.070 & -0.558 & 2.113 &  0.003 & 0.892 & -0.667 & 1.187 &  0.063 \\
 & Our+ST+PCS    & \textbf{0.719} & -0.151 & \textbf{1.289} &  0.518 & 0.579 & -0.162 & 1.037 &  0.523 \\
 & Our+MT+PCS    & 0.752 & \textbf{-0.112} & 1.390 & \textbf{0.568} & 0.582 & \textbf{-0.139} & 0.985 &  0.557 \\
\bottomrule
\end{tabular}
}
\end{table}

\bibliographystyle{unsrtnat}
\bibliography{references}

\section{Appendix / supplemental material}

\subsection{Gradient Backpropagation to Initial State $h_0$.}

To incorporate individualized physiological priors, we initialize the state-space recurrence with a patient-specific hidden state \( h_0 \), generated from historical laboratory values and a static patient embedding. Formally, this initialization is defined as
\begin{equation}
    h_0 = g(e_{\text{patient}}, \ell_{\text{hist}}),
\end{equation}

where \( e_{\text{patient}} \) is the patient embedding, \( \ell_{\text{hist}} \) represents historical lab values, and \( g(\cdot) \) is a learnable function implemented via FiLM modulation and nonlinear projection.

To support gradient-based learning of this initialization, we modify the CUDA-based selective scan kernel to compute gradients with respect to \( h_0 \).

The hidden state evolves as:
\begin{equation}
h_{t+1} = \bar{\mathbf{A}}_t h_t + \bar{\mathbf{B}}_t x_t,
\end{equation}
\begin{equation}
y_t = \bar{\mathbf{C}}_t h_t + \bar{\mathbf{D}} x_t,
\end{equation}
where $\bar{\mathbf{A}}_t = \exp(\Delta_t \mathbf{A})$ is the time-varying state transition matrix after discretization, and $\bar{\mathbf{B}}_t = \bar{\mathbf{B}}_t(x_t)$, $\bar{\mathbf{C}}_t = \bar{\mathbf{C}}_t(x_t)$ are input-dependent.

To compute the gradient of the loss $\mathcal{L}$ with respect to the initial hidden state $h_0$, we unroll the recurrence and apply the chain rule. The influence of $h_0$ on each output $y_t$ propagates through the recurrent Jacobians:

\begin{equation}
\frac{\partial \mathcal{L}}{\partial h_0} = \sum_{t=0}^{L-1} \left( \frac{\partial \mathcal{L}}{\partial y_t} \cdot \bar{\mathbf{C}}_t \cdot \prod_{k=0}^{t-1} \bar{\mathbf{A}}_k \right).
\end{equation}

Here, $\bar{\mathbf{C}}_t$ directly maps hidden states to outputs and the product of $\bar{\mathbf{A}}_k$ captures how $h_0$ influences $h_t$. This formulation allows backpropagation into the patient-specific initialization and supports gradient-based personalization within the Mamba architecture. (This derivation was inspired by a community discussion in the official Mamba repository issue \#285 \#488)


\end{document}